\documentclass[journal]{IEEEtran}
\pdfoutput=1

\usepackage{amsmath,amsfonts,amssymb}
\usepackage{dsfont}  
\usepackage{algorithmic}
\usepackage{algorithm}
\usepackage{array}
\usepackage[caption=false,font=normalsize,labelfont=sf,textfont=sf]{subfig}
\usepackage{textcomp}
\usepackage{stfloats}
\usepackage{url}
\usepackage{verbatim}
\usepackage{graphicx}
\usepackage{cite}
\usepackage{booktabs}
\usepackage{multirow}
\usepackage{xcolor}
\usepackage{colortbl}  
\usepackage{listings}
\lstdefinestyle{promptstyle}{
  basicstyle=\ttfamily\scriptsize,
  breaklines=true,
  breakindent=6pt,
  breakautoindent=true,
  columns=fullflexible,
  keepspaces=true,
  showstringspaces=false,
  xleftmargin=4pt,
  frame=none,
  aboveskip=2pt,
  belowskip=2pt
}

\graphicspath{{figures/}}

\newcommand{\method}{CFCamo}

\newcommand{\benchmark}{CF-COD}

\begin{document}

\title{\method{}: A Counterfactual Detect-or-Abstain Framework for
       Camouflaged Object Detection}

\author{Suhang Li, Osamu Yoshie, and Yuya Ieiri%
\thanks{The authors are with the Graduate School of Information,
Production and Systems, Waseda University, Fukuoka, Japan.}%
\thanks{Corresponding author: Osamu Yoshie (e-mail: yoshie@waseda.jp).}%
}

\markboth{Preprint, 2026}%
{Li \MakeLowercase{\textit{et al.}}: \method{}: A Counterfactual Detect-or-Abstain Framework for COD}

\maketitle

\bstctlcite{IEEEexample:BSTcontrol}

\begin{abstract}
Vision-language reinforcement learning has recently shown strong
target-present localization for camouflaged object detection (COD).
Yet localization is only one side of the decision: when the agent faces
an ordinary image with no camouflaged target, will it still claim that a
camouflaged object exists?
Standard COD training and evaluation data are positive-only, so agents
optimized under this setting can acquire an over-detect bias, a
task-specific form of object hallucination that standard COD evaluation
leaves unmeasured. To quantify this target-absent
behavior, we construct Counterfactual COD (\benchmark{}), a paired
benchmark that removes the camouflaged target from each held-out COD
evaluation image while preserving a plausible background. \benchmark{}
evaluates whether a model detects the target on the original image and abstains on the
target-absent counterfactual, summarized by Pair Accuracy (PA). We further
introduce \method{}, a paired counterfactual framework for COD with
abstention. For training, \method{} optimizes a Qwen3-VL-4B-Instruct
agent with Counterfactual Sequence Policy Optimization (CSPO), which
samples paired original-counterfactual rollouts and uses a
Counterfactual Paired Reward (CPR) to couple original-image detection
with counterfactual abstention. On CAMO-test, \method{} improves
$S_\alpha$ by $+3.7$~pp over the prior RL-based COD baseline; across
\benchmark{}, it reaches $80.0$--$90.8\%$ PA. Ablations show that
removing counterfactual coupling reduces PA to $1.4$--$5.2\%$ despite
strong target-present COD scores, showing that target-present evaluation
alone does not characterize detect-or-abstain behavior. Overall, these results indicate that
\method{} improves COD agents by coupling target-present detection with
target-absent abstention, rather than merely strengthening target-present
localization.
Code and data are available at \url{https://github.com/suhang2000/CFCamo}.
\end{abstract}


\section{Introduction}
\label{sec:intro}

\IEEEPARstart{C}{amouflage} is a widespread biological survival
strategy in which organisms evolve appearance that blends into their
surroundings to evade
detection~\cite{stevens2009Animalcamouflagecurrent,cott1940Adaptivecolorationanimals}.
Camouflaged object detection
(COD)~\cite{fan2020CamouflagedObjectDetectiona,fan2022ConcealedObjectDetection}
aims to localize and segment such targets that visually merge with
complex backgrounds. Unlike salient object detection on visually
prominent targets, COD must operate when the target's local appearance
contrast with the background is minimized. Related
low-contrast segmentation settings arise in medical image analysis
(e.g., polyp segmentation~\cite{fan2020PraNetParallelReverse}),
biological and ecological studies~\cite{perez-delafuente2012Earlyevolutionecology},
and industrial visual inspection
(e.g., surface-defect detection~\cite{tabernik2020Segmentationbaseddeeplearningapproach}).
Early COD systems rely on dense-prediction networks supervised by pixel
masks~\cite{mei2021CamouflagedObjectSegmentationa,sun2022BoundaryGuidedCamouflagedObjecta,hu2023Highresolutioniterativefeedback,pang2024ZoomNeXtUnifiedCollaborative}.
A recent line of work reformulates COD as visual grounding for a
vision-language model (VLM): the model emits spatial prompts and a
frozen SAM-style decoder~\cite{kirillov2023SegmentAnything,ravi2024SAM2Segment}
turns them into a mask. Seg-R1~\cite{you2025SegR1SegmentationCan}
further trains such a VLM by reinforcement learning (RL) with a
mask-level reward and reports strong COD scores.

\begin{figure*}[!t]
  \centering
  \includegraphics[width=\textwidth]{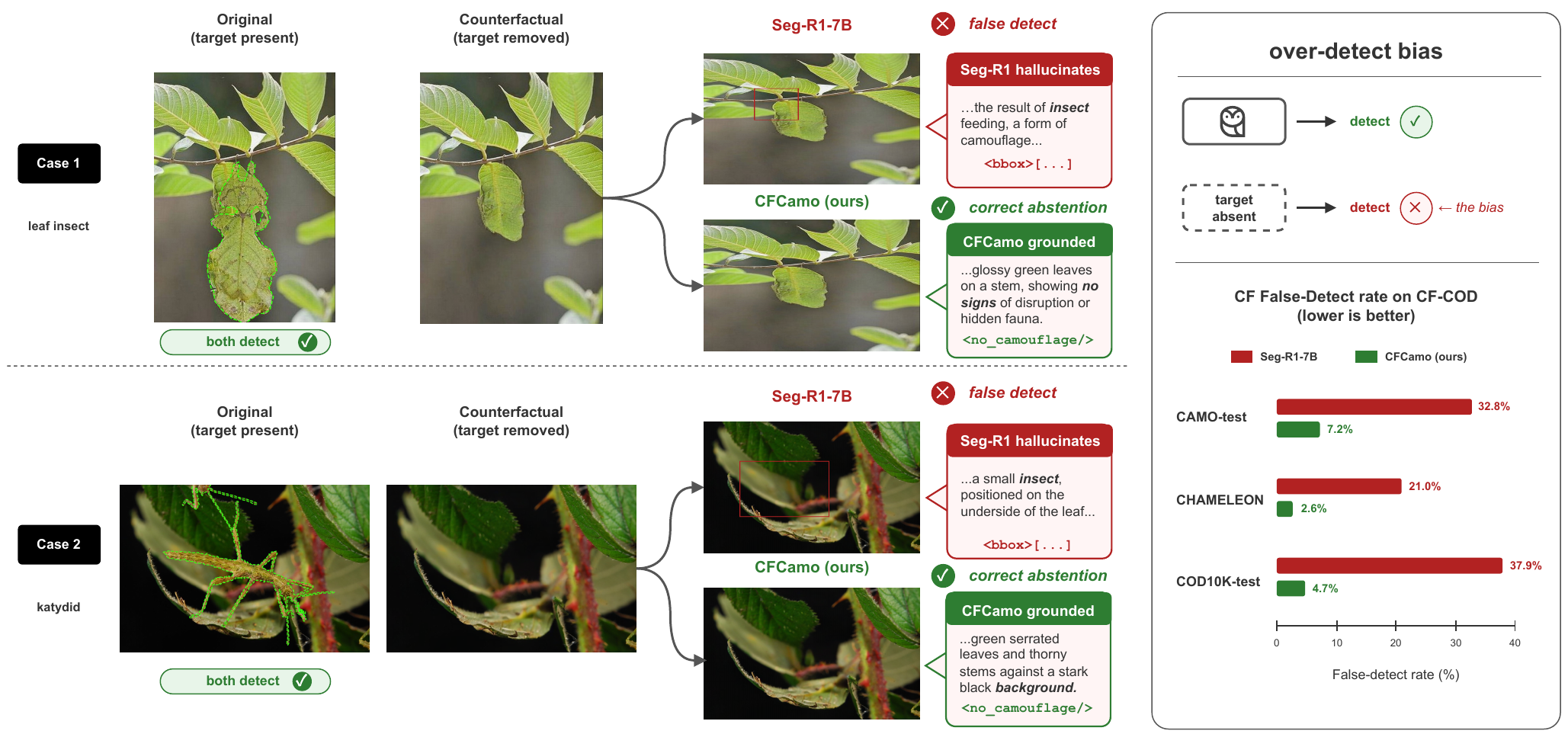}
  \caption{\normalfont Motivation of \method{} and \benchmark{}.
  Standard COD benchmarks evaluate target-present images, whereas
  deployment also contains target-absent scenes. In each paired
  (original, target-absent) example, Seg-R1-7B still predicts a box on
  the target-absent counterfactual, whereas \method{} detects the
  original target and correctly abstains. Right:
  false-detect rates on \benchmark{} counterfactuals (lower is better).}
  \label{fig:motivation}
\end{figure*}

As illustrated in Fig.~\ref{fig:motivation}, strong target-present
accuracy can hide a deployment-relevant failure mode: \emph{over-detect
bias}, a spatial form of object hallucination in which the agent claims a
camouflaged target where none is present. In real-world applications, many
inputs are ordinary target-absent scenes, and repeated false alarms on
such backgrounds reduce deployment reliability. However, standard COD
training and evaluation benchmarks are positive-only: each image is
assumed to contain at least one camouflaged target. A VLM trained under this
distribution can therefore learn a prior of issuing a detection whenever
it receives a COD prompt, while standard COD evaluation does not test the
target-absent case. For example, on target-absent counterfactuals
synthesized by ObjectClear inpainting, even with an explicit abstention
token in the prompt, our re-run of
Seg-R1-7B~\cite{you2025SegR1SegmentationCan} still emits a bounding box
on $38\%$ of COD10K-test images
(Fig.~\ref{fig:motivation}). This behavior indicates target-present predictions in the
absence of target evidence, rather than decisions conditioned only on
what remains visible in the scene.

To address over-detect bias, we propose \method{},
a paired detect-or-abstain COD agent optimized using
Counterfactual Sequence Policy Optimization (CSPO).
Instead of optimizing target-present inputs in isolation,
CSPO operates on original-counterfactual image pairs and enforces a
coupled Counterfactual Paired Reward (CPR) that scores the joint decision
rather than the two scenarios independently. This formulation couples
positive detection with negative abstention in a single RL objective, so
indiscriminate detection no longer gives high reward. To systematically
diagnose this capability, we construct Counterfactual COD (\benchmark{}),
a paired diagnostic benchmark generated by a frozen inpainting
model~\cite{zhao2026PreciseObjectEffect}. Each evaluation image is paired
with its target-absent counterfactual, and Pair Accuracy (PA) measures
whether the model detects on the original image and abstains on the
counterfactual. Trained on a Qwen3-VL-4B-Instruct~\cite{bai2025Qwen3VLTechnicalReport}
backbone, \method{} substantially reduces over-detection on \benchmark{},
improves standard COD performance against task-generic prompt baselines,
and keeps general multimodal capability close to the base model.

Our contributions are as follows.
\begin{itemize}
\item We propose Counterfactual Sequence Policy Optimization (CSPO), a
paired-counterfactual reinforcement learning framework to mitigate
over-detect bias in vision-language models. By integrating
paired-counterfactual rollouts and a sequence-level importance ratio
under a coupled Counterfactual Paired Reward (CPR), CSPO optimizes a
single policy to jointly detect targets on original images and abstain
on target-absent counterfactuals.
\item We construct the Counterfactual COD (\benchmark{}) benchmark, a
paired evaluation framework for camouflaged object detection that scores
joint correctness on original--counterfactual pairs and makes
over-detect bias directly measurable.
\end{itemize}

\section{Related Work}
\label{sec:related}

\subsection{Camouflaged Object Detection}
\label{sec:related:cod}

Camouflaged object detection (COD) localizes objects whose appearance
is intentionally or naturally blended into the surrounding scene
~\cite{le2019Anabranchnetworkcamouflageda}~\cite{fan2020CamouflagedObjectDetectiona}.
The dominant paradigm trains dedicated dense-prediction networks
under pixel-level supervision, with many mechanisms proposed to
handle the appearance ambiguity of camouflaged targets.
PFNet mines distractor regions to suppress confusing
backgrounds~\cite{mei2021CamouflagedObjectSegmentationa}, and BGNet
uses object boundaries as auxiliary supervision to sharpen target
contours~\cite{sun2022BoundaryGuidedCamouflagedObjecta}. To recover
textures missed by the spatial branch, FGSA-Net introduces
frequency-domain guidance into the spatial adaptation of a
pretrained
backbone~\cite{zhang2025FrequencyGuidedSpatialAdaptation}. HitNet
iteratively refines low-resolution predictions with high-resolution
features to avoid detail
loss~\cite{hu2023Highresolutioniterativefeedback}, and ZoomNeXt
integrates multi-scale feature interactions in a collaborative
pyramid~\cite{pang2024ZoomNeXtUnifiedCollaborative}. UGDNet casts
COD as iterative denoising under an explicit uncertainty
condition~\cite{yang2025UncertaintyGuidedDiffusionModel}.

Another line repurposes the Segment Anything Model
family~\cite{kirillov2023SegmentAnything}~\cite{ravi2024SAM2Segment}
as a mask decoder driven by external prompts. GenSAM removes the
per-image prompt requirement by deriving visual prompts from a
single generic text prompt with cross-modal chain-of-thought
prompting~\cite{hu2024Relaximagespecificprompta}. ProMaC turns a multimodal LLM's
hallucinations into candidate prompts, refined through a prompt-mask
cycle where an inpainter generates background-only contrastive
images to filter co-occurrence hallucinations at inference
time~\cite{hu2024LeveragingHallucinationsReduce}. RDVP-MSD couples
a multimodal stepwise chain of thought for caption disambiguation
with region-constrained, dual-stream prompt sampling for foreground
and
background~\cite{yin2025StepwiseDecompositionDualstream}. Beyond
these training-free pipelines, Seg-R1 trains an RL agent that emits
spatial prompts directly from the
image~\cite{you2025SegR1SegmentationCan}. CFCamo formulates COD
as a paired detect-or-abstain problem, and trains the detector
with a pair-level reward over the original image and its inpainted
counterfactual, so that explicit abstention on target-absent scenes
becomes a learned, first-class output.

\subsection{Vision-Language Models for Pixel-Level Understanding}
\label{sec:related:vlmpix}

A broader line of work extends large multimodal
models (LMMs) to pixel-level output.
LISA~\cite{lai2024LISAReasoningSegmentation} first introduced a
dedicated \texttt{<SEG>} token that aligns an LMM with a mask
decoder. GLaMM extends this recipe to grounded multi-object
conversation with region-level
captioning~\cite{rasheed2024GLaMMPixelGrounding}. PixelLM replaces
the mask decoder with a lightweight pixel-token
codebook~\cite{ren2024PixelLMPixelReasoning}. OMG-LLaVA unifies
image-, object-, and pixel-level understanding in a single
LMM~\cite{zhang2024OMGLLaVABridgingImagelevel}. SESAME teaches an
LMM to verify referent existence before segmenting, enabling
abstention on false-premise referring
queries~\cite{wu2024SeeSaySegment}.

Another line applies reinforcement learning to prompt-based
segmentation, building on Group Relative Policy
Optimization (GRPO)~\cite{shao2024DeepSeekMathPushingLimits}.
Seg-Zero~\cite{liu2025SegZeroReasoningChainGuided} freezes SAM2 and
rewards only the geometric prompts (bounding box and points),
whereas Seg-R1~\cite{you2025SegR1SegmentationCan} additionally
feeds the SAM2-decoded mask back into the reward as a mask-quality
signal. SAM-R1~\cite{huang2025SAMR1LeveragingSAM} tightens this
loop with tiered segmentation-accuracy rewards driven by SAM2;
VisionReasoner~\cite{liu2025VisionReasonerUnifiedReasoningIntegrated}
unifies detection, segmentation, and counting under a
Hungarian-matched multi-object reward; and
LENS~\cite{zhu2026LENSLearningSegment} fine-tunes the MLLM and the
SAM2 mask decoder while keeping the segmentation image encoder fixed.
CFCamo extends this paradigm to
camouflaged object detection: the per-image reward is replaced by a
pair-level reward over an original image and its inpainted
counterfactual, which couples detection and abstention in a single
training objective.

\subsection{Counterfactual Reasoning for Vision-Language Model Hallucination}
\label{sec:related:counterfactual}

A separate body of work studies vision-language model hallucination
through object probing, paired comparisons, or counterfactual inputs.
Several benchmarks target different aspects:
POPE~\cite{li2023EvaluatingObjectHallucination} queries object
existence with polling-style questions,
BEAF~\cite{yebin2025BEAFObservingBEforeAFter} evaluates the model
on paired before-after object removal, and
HaloQuest~\cite{wang2024HaloQuestVisualHallucination} covers
multiple hallucination types. Among mitigation methods,
REVERSE~\cite{wu2025GenerateVerifyReducing} suppresses visual
hallucination through hallucination-verification data and
retrospective resampling, and
PAPO~\cite{wang2025PerceptionAwarePolicyOptimization} adds an
implicit perception KL between policy distributions conditioned on
the original versus a corrupted visual input to enforce visual
grounding.

HalluSegBench~\cite{li2026CounterfactualSegmentationReasoning}
constructs a paired benchmark for pixel-grounding hallucination
under counterfactual object substitution.
CIPHER~\cite{dastmalchi2026FightingHallucinationsCounterfactuals}
suppresses vision-induced LVLM hallucination by projecting hidden
states away from an offline hallucination subspace learned from
diffusion-edited counterfactuals, all at inference time. Visual
Jenga~\cite{bhattad2025VisualJengaDiscovering} uses counterfactual
inpainting in a training-free manner to quantify pairwise object
dependencies.
CFCamo applies counterfactual inpainting at training time for
camouflaged object detection, exploiting a task-specific property:
a target-absent counterpart of a camouflaged scene is simply a
natural background image, easier to acquire than the camouflaged scene
itself. We pair such counterparts with the originals and supervise
them with a pair-level reinforcement-learning reward, so the model
learns both to localize camouflaged targets and to abstain on
natural scenes.

\section{Method}
\label{sec:method}

\begin{figure*}[!t]
  \centering
  \includegraphics[width=\textwidth]{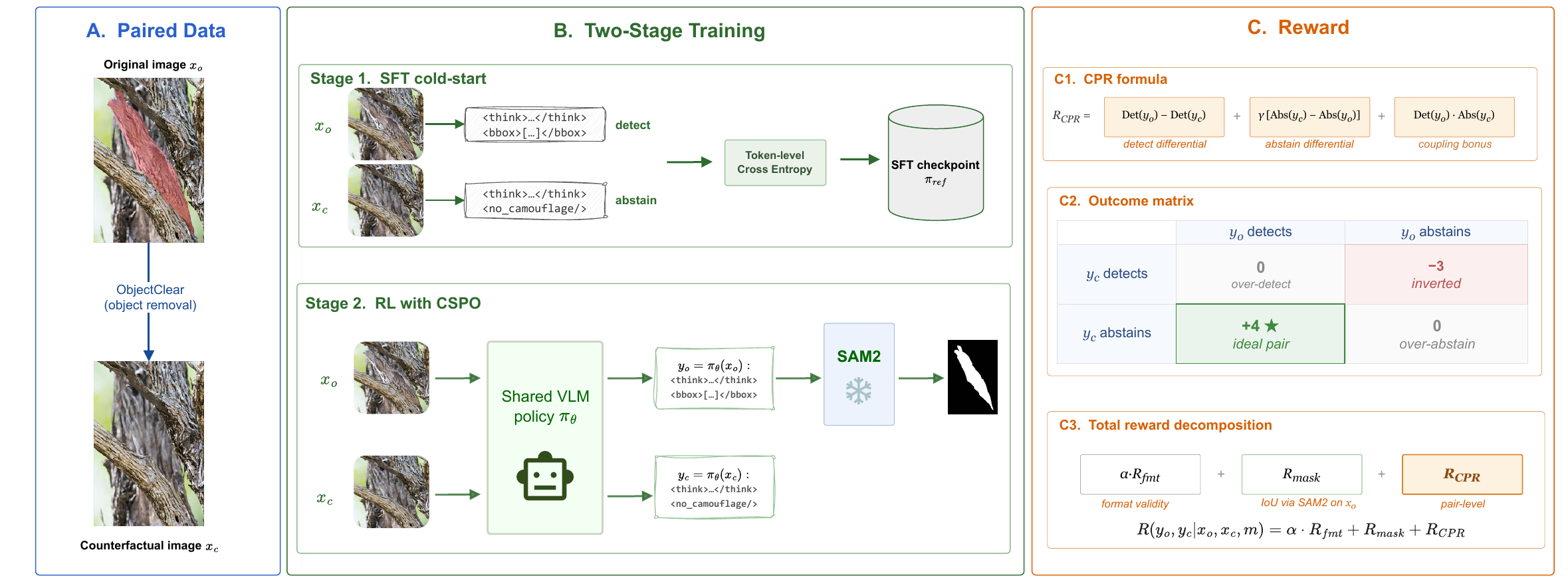}
  \caption{\normalfont Overview of the \method{} framework. Paired training
  data $(x_o, x_c)$ are formed by pairing $x_o$ with a target-absent
  counterpart $x_c$ from an off-the-shelf inpainter. Stage 1 cold-starts
  the VLM with a balanced supervised fine-tuning (SFT) corpus to teach
  the detect-or-abstain format.
  Stage 2 trains the VLM with CSPO; a frozen SAM2 decoder turns
  predicted boxes into masks for the reward's mask-IoU term.}
  \label{fig:framework}
\end{figure*}

\noindent \method{} reframes camouflaged object detection from a
positive-only localization problem into a paired detect-or-abstain
problem (Fig.~\ref{fig:framework}). Standard COD pipelines assume that a camouflaged object is
always present, so a vision-language detector earns reward simply by
emitting a box on every input. This assumption does not hold on
counterfactual scenes that contain no camouflaged object even though
the background remains plausible: an empty scene should not receive a
detection box, otherwise the policy defaults to an over-detect
strategy. \method{} addresses this by optimizing a single VLM policy to
localize on original COD images and to abstain on their inpainted
counterfactuals, within one output space.

Concretely, \method{} extends Group Sequence Policy Optimization
(GSPO)~\cite{zheng2025GroupSequencePolicy}
along three axes, which together form Counterfactual Sequence Policy
Optimization (CSPO): (i)~a paired counterfactual rollout that draws $G$
original and $G$ counterfactual responses per pair
(Section~\ref{sec:method:data}); (ii)~a Counterfactual Paired Reward
(CPR) that couples original-image detection with counterfactual
abstention via detection and abstention indicators and a pair-level
coupling bonus (Section~\ref{sec:method:reward}); and (iii)~a paired sequence ratio that extends GSPO's
length-normalized importance ratio jointly across each
original--counterfactual pair (Section~\ref{sec:method:training}).

\subsection{Problem Formulation}
\label{sec:method:problem}

In the VLM-to-SAM formulation used here, a policy
$\pi_\theta : x \mapsto y$ emits a structured response whose terminal
decision is converted into a mask by SAM2. For a target-present COD
image $x$ with ground-truth mask $m$, the desired terminal decision is a
bounding box for the camouflaged object. \method{} extends this setting to
paired inputs $(x_o, x_c)$, where $x_o$ is an original COD image and
$x_c$ is its target-absent counterfactual counterpart. Writing $y_o = \pi_\theta(x_o)$ and
$y_c = \pi_\theta(x_c)$, the desired behavior is asymmetric across the pair:
\begin{align}
y_o &\rightarrow \text{detect (bbox)}, \notag\\
y_c &\rightarrow \text{abstain}.
\label{eq:paired_task}
\end{align}
Both responses begin with a \texttt{<think>...</think>} reasoning block.
The detect output terminates in one or more bounding boxes, whereas the
abstaining output terminates in the dedicated token
$\langle\texttt{no\_camouflage/}\rangle$.

\subsection{Paired Training Data}
\label{sec:method:data}

\begin{figure}[!t]
  \centering
  \includegraphics[width=\columnwidth]{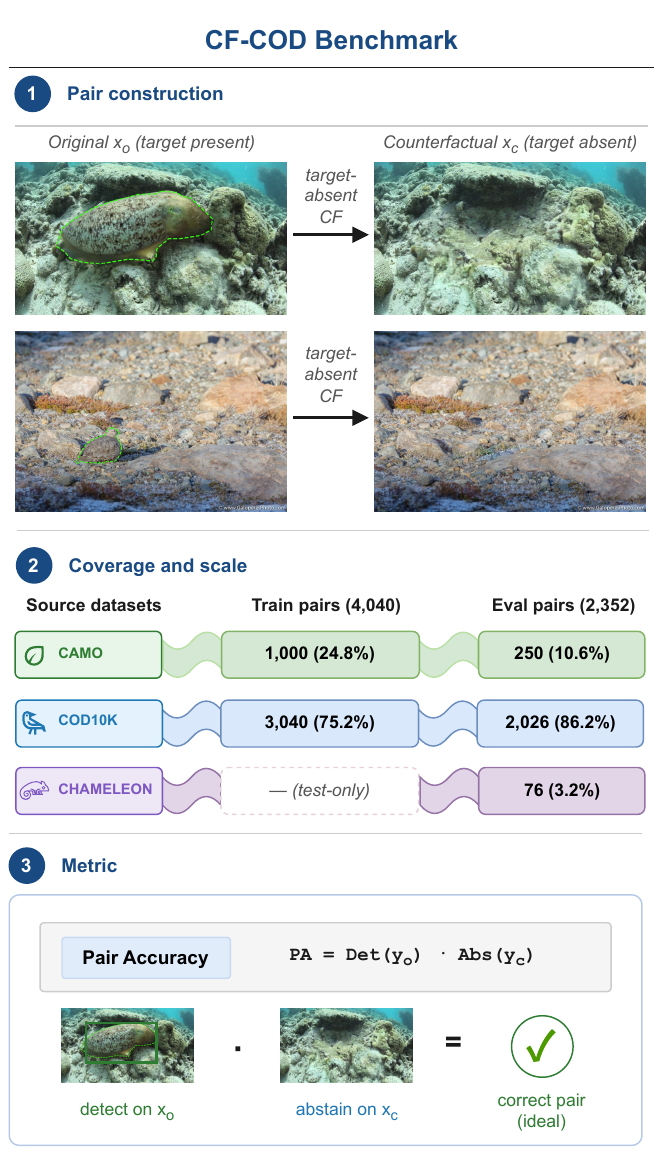}
  \caption{\normalfont Overview of the paired counterfactual data.
  Top: representative paired examples (original $x_o$ with the target outlined
  vs.\ target-absent counterfactual $x_c$). Middle: train/eval split
  ($4{,}040$ training pairs from CAMO/COD10K and
  $2{,}352$ held-out evaluation pairs forming \benchmark{}, with
  CHAMELEON as test-only). Bottom: the Pair Accuracy scoring rule.}
  \label{fig:cf_cod_overview}
\end{figure}

\noindent\textbf{Counterfactual generation.}
We generate counterfactual pairs for both training and evaluation: the
CAMO/COD10K training pairs are used for SFT and CSPO, while
\benchmark{} denotes the held-out paired evaluation suite.
For training, we build the paired training set from the combined CAMO~\cite{le2019Anabranchnetworkcamouflageda}
and COD10K~\cite{fan2020CamouflagedObjectDetectiona} training splits
($4040$ original images). For each original $x_o$,
ObjectClear~\cite{zhao2026PreciseObjectEffect} fills the ground-truth mask
region to produce a target-absent counterpart $x_c$. Compared with random
negative sampling, this construction preserves background, illumination, and
local texture, so abstention must be driven by the disappearance of the
camouflaged evidence rather than dataset-level shortcuts.

\noindent\textbf{Cold-start supervised data.}
Before reinforcement learning, we run a short SFT stage
to teach the model the output schema and a basic reasoning style.
We randomly sample $750$ detect and $750$ abstain examples to form the SFT
pool; rationales for both response types are generated by
an external language model (Gemini~3.1-flash-lite~\cite{googledeepmind2026gemini31flashlite})
at temperature $0$. The main setting uses the balanced $500/500$ subset of
this pool; other ratios are studied in Section~\ref{sec:analysis:sft}.

\noindent\textbf{Reinforcement-learning data.}
The paired training set contains $4040$ pairs in total, yielding $8080$ rollout
inputs per epoch (each pair contributes one rollout on $x_o$ and one on
$x_c$). Our main model is trained for half an epoch, which covers roughly
$2020$ pairs ($\approx 4040$ image-views), matching the size of the
standard COD training set.

\subsection{Reward Design}
\label{sec:method:reward}

The total reward over a paired sample $(x_o, x_c)$ has three
components: a mask-quality reward $R_{\mathrm{mask}}$, a
format-validity reward $R_{\mathrm{fmt}}$ with a small weight
$\alpha=0.1$, and our Counterfactual Paired Reward
$R_{\mathrm{CPR}}$,
\begin{equation}
\label{eq:total_reward}
R(y_o, y_c \mid x_o, x_c, m) =
  \alpha\,R_{\mathrm{fmt}} + R_{\mathrm{mask}} + R_{\mathrm{CPR}}.
\end{equation}
The first two terms follow prior RL-based segmentation
methods~\cite{you2025SegR1SegmentationCan,liu2025SegZeroReasoningChainGuided};
$R_{\mathrm{CPR}}$ ties detection on $x_o$ to abstention on $x_c$
in one reward, so no separate abstention head is required.

\noindent\textbf{Format and mask-quality rewards.}
The format indicator $\mathrm{Fmt}(y)=1$ if $y$ has a well-formed
reasoning block and ends with exactly one valid terminal decision
(a bounding box or the abstention token
$\langle\texttt{no\_camouflage/}\rangle$); $\mathrm{Fmt}(y)=0$
otherwise. The format reward sums $\mathrm{Fmt}$ over the two
responses,
\begin{equation}
\label{eq:r_fmt}
R_{\mathrm{fmt}}(y_o, y_c) = \mathrm{Fmt}(y_o) + \mathrm{Fmt}(y_c).
\end{equation}
For the mask reward, we feed the predicted box into
SAM2~\cite{ravi2024SAM2Segment} and score the resulting mask
against the ground truth $m$ on the original image. Letting
$\mathrm{Det}(y)=1$ if $y$ contains a valid bounding box (and
$0$ otherwise) and
$\mathrm{IoU}_o = \mathrm{IoU}\!\bigl(\mathrm{SAM2}(\mathrm{bbox}(y_o), x_o),\, m\bigr)$,
\begin{equation}
\label{eq:r_mask}
R_{\mathrm{mask}}(y_o, x_o, m) = \mathrm{IoU}_o \cdot \mathrm{Det}(y_o),
\end{equation}
which is zero whenever $y_o$ abstains or is malformed.

\noindent\textbf{Counterfactual paired reward.}
On the counterfactual side, the abstention indicator
$\mathrm{Abs}(y)=1$ if $y$ ends with the abstention token (and $0$
otherwise). $R_{\mathrm{CPR}}$ then combines the four pair-level
outcomes antisymmetrically through $\mathrm{Det}$ and
$\mathrm{Abs}$:
\begin{equation}
\label{eq:cpr}
\begin{aligned}
R_{\mathrm{CPR}}(y_o, y_c) =\;& \phantom{+}\mathrm{Det}(y_o) - \mathrm{Det}(y_c) && \text{\;(detect)} \\
 &+ \gamma\bigl[\mathrm{Abs}(y_c) - \mathrm{Abs}(y_o)\bigr] && \text{\;(abstain)} \\
 &+ \mathrm{Det}(y_o)\,\mathrm{Abs}(y_c). && \text{\;(coupling)}
\end{aligned}
\end{equation}
Each differential rewards the correct behavior on its side and
penalizes it on the other: the over-detect bias appears as the
$-\mathrm{Det}(y_c)$ term, and over-abstention as
$-\mathrm{Abs}(y_o)$. The coupling bonus fires only on the ideal
pair, adding a pair-level signal that per-image rewards alone
cannot deliver.

We use $\gamma=2$ throughout. By antisymmetry, the two trivial
strategies receive zero CPR: the detect differential
$\mathrm{Det}(y_o)-\mathrm{Det}(y_c)$ vanishes under over-detection,
and the abstain differential
$\gamma[\mathrm{Abs}(y_c)-\mathrm{Abs}(y_o)]$ vanishes under
over-abstention. The ideal pair receives the largest CPR value,
$1+\gamma+1=4$, whereas the inverted abstain-detect pair receives
$-1-\gamma=-3$.

\subsection{Two-Stage Training}
\label{sec:method:training}

\noindent\textbf{Stage 1: Supervised fine-tuning.}
We first fine-tune the base VLM on the cold-start data
$\mathcal{D}_{\mathrm{SFT}}$ of Section~\ref{sec:method:data}
by minimizing the token-level cross-entropy
\begin{equation}
\label{eq:sft}
\mathcal{L}_{\mathrm{SFT}}(\theta) = -\mathbb{E}_{(x,\,y^{*})\sim\mathcal{D}_{\mathrm{SFT}}}
\sum_{t=1}^{|y^{*}|} \log \pi_\theta\!\left(y^{*}_{t} \mid y^{*}_{<t},\, x\right),
\end{equation}
where $y^{*}$ is the supervised response (a reasoning block followed by a
bounding box or the abstention token). This teaches the model the
\method{} response format before any reward is applied; the resulting
checkpoint then initializes Stage~2 and serves as the reference policy
$\pi_{\mathrm{ref}}$ for the KL anchor in Eq.~\eqref{eq:rl_objective}.

\noindent\textbf{Stage 2: Reinforcement learning.}
We then optimize the reward of Section~\ref{sec:method:reward} with
CSPO, which follows the sequence-level clipped objective of
GSPO~\cite{zheng2025GroupSequencePolicy}, itself a sequence-level
variant of GRPO~\cite{shao2024DeepSeekMathPushingLimits}. For
each pair $(x_o, x_c)$ we sample $G$ rollouts per side from the
behavior policy $\pi_{\theta_{\mathrm{old}}}$, pair them by
index as $o_i = (y_{o,i}, y_{c,i})$, and score each pair with
the total reward $R(o_i)$ from Eq.~\eqref{eq:total_reward}. The
policy maximizes
\begin{equation}
\label{eq:rl_objective}
\mathcal{J}_{\mathrm{CSPO}}(\theta) = \mathbb{E}\!\left[
\begin{aligned}
&\frac{1}{G}\sum_{i=1}^{G}
\min\!\bigl(s_i \hat{A}_i,\,\mathrm{clip}(s_i,1{-}\epsilon,1{+}\epsilon)\hat{A}_i\bigr)\\
&\hspace{2cm}- \beta_{\mathrm{kl}}\,D_{\mathrm{KL}}\!\left(\pi_\theta\,\|\,\pi_{\mathrm{ref}}\right)
\end{aligned}
\right],
\end{equation}
where the sequence-level ratio $s_i$ extends GSPO to the matched
pair as the length-normalized geometric mean of the joint
likelihood ratio,
\begin{equation}
\label{eq:gspo_ratio}
s_i = \left(
\frac{\pi_\theta(y_{o,i}\mid x_o)\,\pi_\theta(y_{c,i}\mid x_c)}
     {\pi_{\theta_{\mathrm{old}}}(y_{o,i}\mid x_o)\,\pi_{\theta_{\mathrm{old}}}(y_{c,i}\mid x_c)}
\right)^{\!1 / (|y_{o,i}| + |y_{c,i}|)},
\end{equation}
which stabilizes training on variable-length reasoning
trajectories. $\hat{A}_i$ is the group-relative advantage
normalized within $\{R(o_1),\dots,R(o_G)\}$, $\pi_{\mathrm{ref}}$
is the SFT checkpoint, and $\epsilon$ and $\beta_{\mathrm{kl}}$
are the clip threshold and KL coefficient.

Our main model fine-tunes all parameters under the half-epoch
budget of Section~\ref{sec:method:data}. We also train a
low-rank adaptation (LoRA) variant as a compute-efficient
alternative, used in our ablations.

\section{Experiments}
\label{sec:results}

\subsection{Experimental Setup}
\label{sec:results:setup}

\noindent\textbf{Datasets.}
Training uses the CAMO~\cite{le2019Anabranchnetworkcamouflageda} (1{,}000 images)
and COD10K~\cite{fan2020CamouflagedObjectDetectiona} (3{,}040 images) train splits,
totaling 4{,}040 images, each paired with an ObjectClear-inpainted,
target-absent counterfactual.
Testing covers three benchmarks: CAMO-test (250 images),
CHAMELEON~\cite{skurowski2018animal} (76 images), and COD10K-test (2{,}026 images);
their held-out original--counterfactual pairs constitute \benchmark{}.

\noindent\textbf{Standard COD metrics.}
Standard COD performance uses four metrics:
(1) structure-measure $S_\alpha$~\cite{fan2017StructureMeasureNewWay},
which captures structural similarity between prediction and ground truth in
region-aware and object-aware ways;
(2) enhanced-alignment measure $E_\phi$~\cite{fan2018Enhancedalignmentmeasurebinary},
which combines local pixel alignment with global statistics, with the mean score
reported;
(3) weighted F-measure $F_\beta^w$~\cite{margolin2014HowEvaluateForeground},
which weights precision and recall toward perceptually significant errors; and
(4) mean absolute error $M$~\cite{perazzi2012SaliencyfiltersContrast},
the average pixel-wise difference between the predicted and ground-truth masks.
Higher is better for $S_\alpha$, $E_\phi$, and $F_\beta^w$; lower is better for $M$.

\noindent\textbf{\benchmark{} paired protocol.}
For \benchmark{} pairs, using the per-sample predicates $\mathrm{Det}$ and
$\mathrm{Abs}$ defined in Section~\ref{sec:method:reward}, we report Detection
Rate $D_o = \tfrac{1}{N}\sum_{i=1}^{N}\mathrm{Det}(y_o^{(i)})$ on original COD images
and Abstention Rate $A_c = \tfrac{1}{N}\sum_{i=1}^{N}\mathrm{Abs}(y_c^{(i)})$ on
counterfactuals.
The headline metric is the joint Pair Accuracy
\begin{equation}
\mathrm{PA} = \frac{1}{N}\sum_{i=1}^{N} \mathrm{Det}(y_o^{(i)}) \cdot \mathrm{Abs}(y_c^{(i)}),
\label{eq:pa}
\end{equation}

\noindent\textbf{Cold-start data.}
SFT uses 500 detection examples (original image, ground-truth bounding box,
and a rationale generated by
Gemini~3.1 Flash-Lite~\cite{googledeepmind2026gemini31flashlite} at temperature~0)
and 500 abstention examples on the ObjectClear counterfactuals,
giving a balanced 1{,}000-example corpus.
Section~\ref{sec:analysis:sft} examines the detect-to-abstain ratio.

\noindent\textbf{Training.}
For the same-backbone comparison, the base, SFT, and \method{} variants build
on Qwen3-VL-4B-Instruct~\cite{bai2025Qwen3VLTechnicalReport}.
The Qwen3-VL processor caps images at $768^2$ pixels, and bounding boxes use
the model's native $[0,1000]$ coordinate range.
SFT runs for one epoch with AdamW (lr $2{\times}10^{-5}$, effective batch~16).
For RL, CSPO (Section~\ref{sec:method:training}) optimizes the reward of
Section~\ref{sec:method:reward} for half an epoch, with rollout group $G{=}8$,
clip $\epsilon{=}0.2$, sampling temperature~1.0 and top-$p$~0.9, and greedy
decoding at evaluation.
The Full~FT variant uses AdamW (lr $10^{-6}$, weight decay $10^{-2}$,
global batch~32, $\beta_{\mathrm{kl}}{=}5{\times}10^{-2}$) on
$4{\times}$ NVIDIA A800 (80~GB each).
The LoRA variant uses rank~64, $\alpha_{\mathrm{lora}}{=}128$,
lr $5{\times}10^{-6}$, batch~16, and $\beta_{\mathrm{kl}}{=}10^{-2}$, and
trains on a single NVIDIA RTX PRO~6000 (96~GB) as a compute-efficient
alternative.
At inference, a frozen SAM2.1-hiera-large~\cite{ravi2024SAM2Segment} decoder
maps predicted boxes to masks, served by
vLLM~\cite{kwon2023EfficientMemoryManagement} with
FlashAttention-2~\cite{dao2023FlashAttention2FasterAttention}.
The standard prompt $P_{\mathrm{std}}$ and the detect-or-abstain prompt
$P_{\mathrm{da}}$ are listed in Appendix~\ref{app:prompts}.

\begin{table*}[!t]
\caption{\normalfont Quantitative comparison on three COD benchmarks.
All listed methods follow the task-generic prompt setting; Seg-R1 and
\method{} are RL-based methods. Best score per metric in \textbf{bold};
our rows are shaded.
\textsuperscript{\dag}Seg-R1 numbers are obtained from our re-run under its native protocol.}
\label{tab:std_cod}
\setlength{\tabcolsep}{4pt}
\resizebox{\textwidth}{!}{%
\footnotesize
\begin{tabular}{lccccccccccccc}
\toprule
\multirow{2}{*}{Method} & \multirow{2}{*}{Venue}
  & \multicolumn{4}{c}{CAMO-test}
  & \multicolumn{4}{c}{CHAMELEON}
  & \multicolumn{4}{c}{COD10K-test} \\
\cmidrule(lr){3-6} \cmidrule(lr){7-10} \cmidrule(lr){11-14}
 &
  & $S_\alpha\!\uparrow$ & $E_\phi\!\uparrow$ & $F_\beta^w\!\uparrow$ & $M\!\downarrow$
  & $S_\alpha\!\uparrow$ & $E_\phi\!\uparrow$ & $F_\beta^w\!\uparrow$ & $M\!\downarrow$
  & $S_\alpha\!\uparrow$ & $E_\phi\!\uparrow$ & $F_\beta^w\!\uparrow$ & $M\!\downarrow$ \\
\midrule
GPT-4V+SAM~\cite{openai2023gpt4v,kirillov2023SegmentAnything} & ICCV'23 & .573 & .666 & .466 & .206  & .637 & .710 & .557 & .180  & .601 & .672 & .448 & .187 \\
SEEM~\cite{zou2023Segmenteverythingeverywhere} & NeurIPS'23 & .404 & .315 & .023 & .192  & .454 & .307 & .011 & .094  & .425 & .280 & .001 & .143 \\
X-Decoder~\cite{zou2023GeneralizedDecodingPixel} & CVPR'23 & .709 & .745 & .628 & .104  & .716 & .748 & .654 & .124  & .652 & .705 & .556 & .171 \\
LLaVA1.5+SAM~\cite{liu2024ImprovedBaselinesVisual,kirillov2023SegmentAnything} & CVPR'24 & .501 & .585 & .401 & .314  & .666 & .718 & .561 & .168  & .662 & .728 & .530 & .170 \\
Grounded SAM~\cite{ren2024GroundedSAMAssembling} & arXiv'24 & .707 & .753 & .656 & .157  & .744 & .776 & .662 & .122  & .764 & .813 & .670 & .085 \\
GenSAM~\cite{hu2024Relaximagespecificprompta} & AAAI'24 & .719 & .775 & .659 & .113  & .764 & .807 & .680 & .090  & .775 & .838 & .681 & .067 \\
ProMaC~\cite{hu2024LeveragingHallucinationsReduce} & NeurIPS'24 & .767 & .846 & .725 & .090  & .833 & .899 & .790 & .044  & .805 & .876 & .716 & .042 \\
CLIP-Surgery+SAM~\cite{li2025closerlookexplainability,kirillov2023SegmentAnything} & PR'25 & .612 & .692 & .520 & .189  & .689 & .741 & .606 & .147  & .629 & .698 & .488 & .173 \\
RDVP-MSD~\cite{yin2025StepwiseDecompositionDualstream} & MM'25 & .796 & .848 & .785 & .081  & .832 & .904 & .814 & .040  & .825 & .877 & .775 & .038 \\
\midrule
Seg-R1-7B~\cite{you2025SegR1SegmentationCan}\textsuperscript{\dag} & NeurIPS-WS'25 & .797 & .843 & .753 & .098 & .846 & .896 & .815 & .083 & .844 & .888 & .775 & .055 \\
\rowcolor{gray!15}
\textbf{\method{}-4B (Full FT)} & Ours
  & \textbf{.834} & \textbf{.884} & \textbf{.807} & \textbf{.059}
  & \textbf{.878} & \textbf{.927} & .841 & \textbf{.033}
  & \textbf{.875} & \textbf{.930} & \textbf{.832} & \textbf{.027} \\
\rowcolor{gray!15}
\textbf{\method{}-4B (LoRA)} & Ours
  & .827 & .882 & .804 & .070
  & \textbf{.878} & .926 & \textbf{.842} & .034
  & \textbf{.875} & .928 & \textbf{.832} & .031 \\
\bottomrule
\end{tabular}%
}
\vspace{2pt}\\
{\footnotesize \textsuperscript{\dag}Seg-R1 originally released checkpoints and per-dataset
numbers for CAMO-test and COD10K-test. For a matched comparison we re-ran its official
checkpoint with the native prompt and $768{\times}768$ image-resize protocol on the same
SAM2.1-hiera-large decoder and metric stack used for our rows.}
\end{table*}

\subsection{Standard COD Benchmark}
\label{sec:results:std}

\noindent\textbf{Quantitative Comparison.}
Table~\ref{tab:std_cod} compares \method{} against prior task-generic prompt COD
methods. For a fair comparison, \method{} is evaluated under the standard prompt
$P_{\mathrm{std}}$, which requires every method to output a bounding box without
any abstention option; invalid or empty predictions are treated as an all-zero
mask when computing metrics. We report two variants: \method{}-4B (Full~FT) trained with full-parameter
RL, and \method{}-4B (LoRA) as a compute-efficient alternative.

The LoRA variant remains close to Full~FT: it trails by $0.7$~pp in
$S_\alpha$ on CAMO-test and is tied on CHAMELEON and COD10K-test, indicating that
low-rank adaptation is competitive under this setting.
Compared with the prior task-generic prompt baseline in
Table~\ref{tab:std_cod}, RDVP-MSD~\cite{yin2025StepwiseDecompositionDualstream},
\method{}-4B (Full~FT) improves $S_\alpha$ consistently, by
$+3.8$ to $+5.0$~pp.
Compared with Seg-R1-7B~\cite{you2025SegR1SegmentationCan}, the closest
RL-based competitor, \method{}-4B (Full~FT) improves $S_\alpha$ by
$+3.1$ to $+3.7$~pp.

\begin{figure*}[!t]
  \centering
  \includegraphics[width=\textwidth]{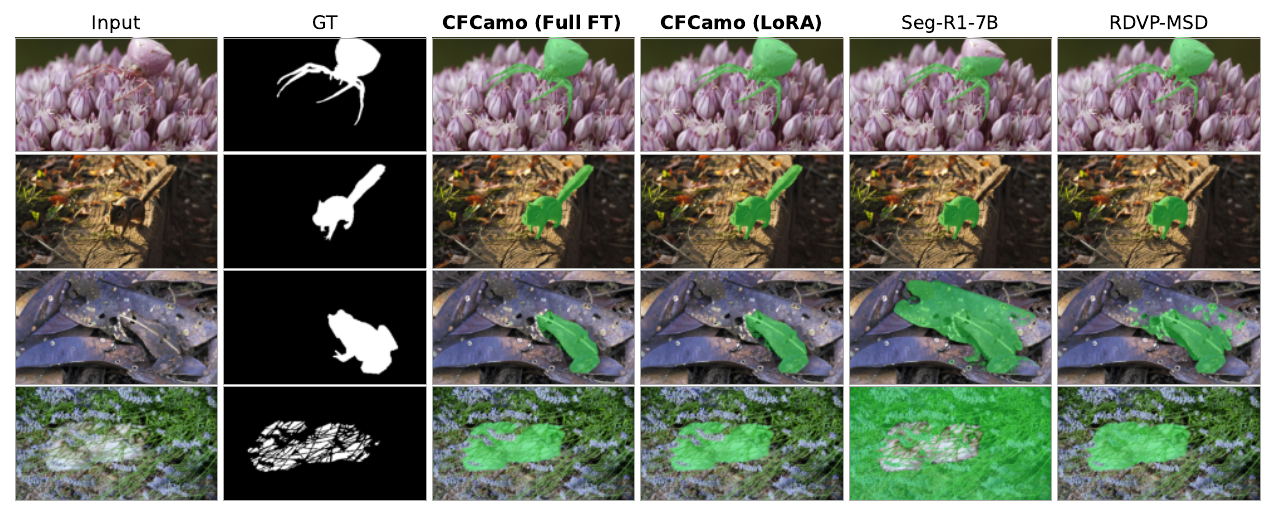}
  \caption{\normalfont Qualitative comparison of \method{} (both variants) against leading task-generic prompt methods.
  Predicted masks are shown as green overlays.}
  \label{fig:qualitative}
\end{figure*}

\noindent\textbf{Qualitative Comparison.}
Fig.~\ref{fig:qualitative} shows a qualitative comparison of \method{} (both variants)
against leading task-generic prompt methods.
In these examples, \method{} produces masks that conform tightly
to the camouflaged target, whereas Seg-R1-7B and RDVP-MSD frequently miss the
object boundary or produce fragmented overlays.
The LoRA outputs are visually close to Full~FT across the four examples,
supporting its role as a compute-efficient alternative.

\begin{table*}[!t]
\caption{\normalfont Comparison on the \benchmark{} paired benchmark.
$D_o$ (\%) is the detection rate on original images and $A_c$ (\%) the
abstention rate on ObjectClear counterfactuals; PA (\%) is the joint Pair
Accuracy of Eq.~\eqref{eq:pa}. Best per metric in bold.}
\label{tab:cf_cod}
\setlength{\tabcolsep}{4pt}
\resizebox{\textwidth}{!}{%
\footnotesize
\begin{tabular}{lcccccccccc}
\toprule
\multirow{2}{*}{Method} & \multirow{2}{*}{Backbone}
  & \multicolumn{3}{c}{CAMO-test}
  & \multicolumn{3}{c}{CHAMELEON}
  & \multicolumn{3}{c}{COD10K-test} \\
\cmidrule(lr){3-5} \cmidrule(lr){6-8} \cmidrule(lr){9-11}
 &
  & $D_o\!\uparrow$ & $A_c\!\uparrow$ & $\mathrm{PA}\!\uparrow$
  & $D_o\!\uparrow$ & $A_c\!\uparrow$ & $\mathrm{PA}\!\uparrow$
  & $D_o\!\uparrow$ & $A_c\!\uparrow$ & $\mathrm{PA}\!\uparrow$ \\
\midrule
Base                       & Qwen3-VL-4B & 68.0 & 92.0 & 60.8 & 88.2 & 97.4 & 85.5 & 73.9 & 88.7 & 64.6 \\
SFT-only ($r{=}1{:}1$)     & Qwen3-VL-4B & 72.0 & 86.4 & 61.2 & 71.1 & \textbf{100.0} & 71.1 & 54.0 & \textbf{96.0} & 50.4 \\
Seg-R1-7B~\cite{you2025SegR1SegmentationCan} & Qwen2.5-VL-7B & 78.4 & 67.2 & 51.6 & 89.5 & 79.0 & 72.4 & 81.2 & 62.1 & 48.5 \\
\midrule
\rowcolor{gray!15}
\textbf{\method{}-4B (Full FT)} & Qwen3-VL-4B & 86.0 & \textbf{92.8} & 80.0 & \textbf{93.4} & 97.4 & \textbf{90.8} & 92.3 & 95.3 & 87.7 \\
\rowcolor{gray!15}
\textbf{\method{}-4B (LoRA)}    & Qwen3-VL-4B & \textbf{88.8} & 92.4 & \textbf{81.6} & \textbf{93.4} & 97.4 & \textbf{90.8} & \textbf{94.0} & 94.2 & \textbf{88.2} \\
\bottomrule
\end{tabular}%
}
\end{table*}

\subsection{CF-COD Paired Benchmark}
\label{sec:results:cf}

\noindent\textbf{Quantitative Comparison.}
The standard COD results in Table~\ref{tab:std_cod} show strong
target-present localization, but they do not test the complementary
target-absent case. A model may still mark ordinary background regions as
camouflaged objects on the counterfactual; such predictions are
hallucinated target-present decisions on target-absent inputs.
Table~\ref{tab:cf_cod}
therefore reports $D_o$, $A_c$, and PA for the Qwen3-VL-4B-Instruct base,
our $1{,}000$-example cold-start SFT, Seg-R1-7B with an added
\texttt{<no\_camouflage/>} option, and both \method{} variants.

Seg-R1-7B shows the over-detection pattern that \benchmark{} is designed
to measure. On COD10K-test, for example, $D_o$ is
$81.2$\% while $A_c$ is only $62.1$\%, yielding $48.5$\% PA; thus, only
about half of the original-counterfactual pairs are judged correctly. The
same direction appears on CAMO-test and CHAMELEON. The Qwen3-VL-4B-Instruct base
shows the opposite asymmetry: on CAMO-test, $D_o$ is $68.0$\% whereas $A_c$ is
$92.0$\%, so its PA mainly reflects over-abstention rather than balanced
paired decisions. The base can emit either \texttt{<bbox>...</bbox>} or
\texttt{<no\_camouflage/>}, but it does not follow the full response schema:
a \texttt{<think>...</think>} reasoning block followed by one of these decision
tokens. Cold-start SFT aligns this schema; however, PA remains near the base
on CAMO-test and drops on COD10K-test, indicating that format alignment alone is
not enough to resolve the paired-decision imbalance
(Section~\ref{sec:analysis:sft}).

Under CSPO, $D_o$ and $A_c$ increase together to $86$--$94$\% and
$92$--$97$\%, respectively. \method{}-4B (Full~FT) reaches $87.7$\% PA on
COD10K-test, gaining $+39.2$~pp over Seg-R1-7B, and $80.0$\% PA on CAMO-test,
$+19.2$~pp over the base. The same-backbone ablation in
Section~\ref{sec:analysis:reward} further separates the method design
from backbone capacity: when the counterfactual coupling is removed, the
same-backbone positive-only RL variant falls to $1.4$--$5.2$\% PA. The LoRA
variant remains comparable to Full~FT, showing that a parameter-efficient
training setting can obtain similar paired-benchmark gains at lower training
cost.

\section{Ablation Study}
\label{sec:analysis}

\subsection{SFT Data Composition}
\label{sec:analysis:sft}

\begin{figure}[!t]
  \centering
  \includegraphics[width=\columnwidth]{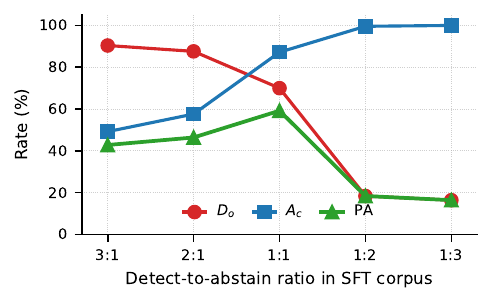}
  \caption{\normalfont SFT detection-abstention trade-off on CAMO-test under
  five detect-to-abstain ratios at fixed corpus size ($1{,}000$).
  Detection rate $D_o$ and abstention rate $A_c$ move in opposite
  directions, and PA peaks at the balanced setting.}
  \label{fig:sft_ratio}
\end{figure}

Figure~\ref{fig:sft_ratio} asks whether the paired-decision asymmetry in
Table~\ref{tab:cf_cod} can be corrected by changing SFT data composition
alone. We keep the SFT corpus size fixed at $1{,}000$ examples and retrain
five models with detect-to-abstain ratios from $3{:}1$ to $1{:}3$.
The balanced model in Table~\ref{tab:cf_cod} is the $1{:}1$ point in this
sweep.

Changing the SFT ratio mainly moves the model along a detection-abstention
trade-off. As the ratio shifts from $3{:}1$ to $1{:}3$, $D_o$
falls from $90.4\%$ to $16.4\%$, while $A_c$ rises from $49.2\%$ to
$100.0\%$. PA therefore forms an inverted-U curve, peaking at $59.2\%$ at
the balanced ratio, still below the $80$--$91\%$ range reached by full
\method{} in Table~\ref{tab:cf_cod}. Format validity remains above $98\%$
for all five ratios, so the cold-start stage has already learned the response
schema. The limiting factor is therefore not schema acquisition, but the
independent SFT objective, which shifts the decision bias toward one side of
the pair. Section~\ref{sec:analysis:reward} therefore examines whether a
paired RL reward can optimize detection and abstention jointly.

\subsection{Reward Component Ablation}
\label{sec:analysis:reward}

\begin{table}[!t]
\caption{\normalfont CF-COD paired ablation of the coupling and
mask-reward components under CSPO, with the same \method{}-4B (LoRA)
training setup as Table~\ref{tab:cf_cod}. $D_o$,
$A_c$, and PA are paired-decision rates (\%). Best values are bolded within
each dataset and metric.}
\label{tab:reward_ablation_pair}
\centering\footnotesize
\setlength{\tabcolsep}{4pt}
\begin{tabular}{llccc}
\toprule
Variant & Dataset & $D_o\!\uparrow$ & $A_c\!\uparrow$ & $\mathrm{PA}\!\uparrow$ \\
\midrule
\multirow{3}{*}{\textbf{\method{}-4B (LoRA)}}
  & CAMO-test   & 88.8 & 92.4 & \textbf{81.6} \\
  & CHAMELEON   & 93.4 & \textbf{97.4} & 90.8 \\
  & COD10K-test & 94.0 & 94.2 & \textbf{88.2} \\
\midrule
\multirow{3}{*}{w/o CF coupling}
  & CAMO-test   & \textbf{99.6} &  5.2 &  5.2 \\
  & CHAMELEON   & \textbf{100.0} &  2.6 &  2.6 \\
  & COD10K-test & \textbf{99.9} &  1.4 &  1.4 \\
\midrule
\multirow{3}{*}{w/o SAM mask}
  & CAMO-test   & 82.8 & \textbf{95.6} & 78.8 \\
  & CHAMELEON   & 94.7 & \textbf{97.4} & \textbf{92.1} \\
  & COD10K-test & 92.2 & \textbf{95.5} & 87.7 \\
\bottomrule
\end{tabular}
\end{table}

\begin{table}[!t]
\caption{\normalfont Standard COD ablation of the coupling and
mask-reward components under CSPO, evaluated with the $P_{\mathrm{std}}$
force-detect prompt, same setup as Table~\ref{tab:reward_ablation_pair}.
Best values are bolded within each dataset and metric.}
\label{tab:reward_ablation_std}
\centering\footnotesize
\setlength{\tabcolsep}{4pt}
\begin{tabular}{llcccc}
\toprule
Variant & Dataset & $S_\alpha\!\uparrow$ & $E_\phi\!\uparrow$ & $F_\beta^w\!\uparrow$ & $M\!\downarrow$ \\
\midrule
\multirow{3}{*}{\textbf{\method{}-4B (LoRA)}}
  & CAMO-test   & \textbf{.827} & \textbf{.882} & \textbf{.804} & .070 \\
  & CHAMELEON   & \textbf{.878} & \textbf{.926} & .842 & \textbf{.034} \\
  & COD10K-test & .875 & .928 & .832 & .031 \\
\midrule
\multirow{3}{*}{w/o CF coupling}
  & CAMO-test   & .826 & .880 & .792 & \textbf{.063} \\
  & CHAMELEON   & .875 & .916 & .840 & .040 \\
  & COD10K-test & \textbf{.879} & \textbf{.932} & \textbf{.833} & \textbf{.023} \\
\midrule
\multirow{3}{*}{w/o SAM mask}
  & CAMO-test   & .814 & .870 & .796 & .085 \\
  & CHAMELEON   & .876 & .920 & \textbf{.845} & .042 \\
  & COD10K-test & .867 & .917 & .822 & .040 \\
\bottomrule
\end{tabular}
\end{table}

To isolate the roles of the counterfactual coupling and the SAM mask reward
within CSPO (Section~\ref{sec:method:training}), we train two ablations with
the same LoRA setup as \method{}-4B (LoRA). The \textbf{w/o CF coupling}
variant removes the counterfactual branch, yielding positive-only RL on
original images. The \textbf{w/o SAM mask} variant keeps paired
counterfactual training but replaces the SAM2-decoded mask IoU with a plain
bounding-box IoU.
Table~\ref{tab:reward_ablation_pair} evaluates paired decisions on
\benchmark{}, and Table~\ref{tab:reward_ablation_std} evaluates the same
checkpoints under the standard COD prompt.

The first ablation shows why target-present COD scores alone are incomplete
for this problem. Without CF coupling, the model almost always detects on natural
images: $D_o$ reaches $99.6$--$100.0\%$. This behavior can look favorable
under the standard force-detect protocol, where every test image contains a
target; indeed, the same checkpoint matches \method{}-4B (LoRA) within
$0.004$ $S_\alpha$ and gives the best COD10K-test standard scores. However,
the same policy does not transfer to counterfactual images: $A_c$ remains at
$1.4$--$5.2\%$, keeping PA low. In other words, the strong
standard-COD numbers partly reflect over-detection rather than balanced
paired reasoning.

The SAM mask reward affects a different part of the behavior. Removing it
keeps PA within $2.8$~pp of \method{}-4B (LoRA), but lowers standard mask
quality, with $S_\alpha$ lower by $0.013$ on CAMO-test and $0.008$ on
COD10K-test. Thus, CF coupling supplies the target-absent constraint needed
for paired decision behavior, whereas the SAM mask reward mainly improves
the segmentation quality of detected objects.

\subsection{Training Trajectory and Schema Stability}
\label{sec:analysis:trajectory}

\begin{figure}[!t]
  \centering
  \includegraphics[width=\columnwidth]{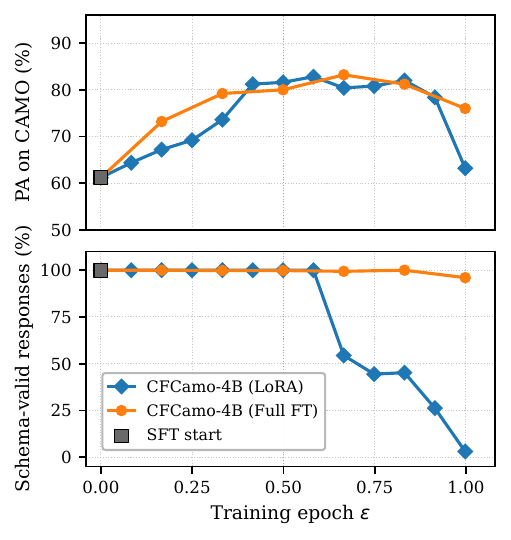}
  \caption{\normalfont Training trajectory on CAMO-test under $P_{\mathrm{da}}$.
  Top: PA over training epoch~$\varepsilon$; $\varepsilon{=}0.5$
  corresponds to the half-epoch budget ($4040$ image-views), and
  $\varepsilon{=}1.0$ to one epoch ($8080$ image-views). Bottom:
  schema-valid response rate, measuring whether the output keeps the required
  reasoning block and terminal decision token.}
  \label{fig:policy_collapse}
\end{figure}

Figure~\ref{fig:policy_collapse} traces CAMO-test PA and schema-valid response
rate along the RL trajectory. At the half-epoch budget, the LoRA and
Full~FT variants reach similar CAMO-test PA ($81.6\%$ and $80.0\%$,
respectively), while both retain fully valid output schemas. This supports
the matched-budget comparison used for the main results: low-rank adaptation
can reach the same performance regime as full-parameter training before
longer optimization changes the response format.

Continuing training reveals the stability difference. The LoRA trajectory
remains near its PA peak shortly after $\varepsilon{=}0.5$, but its
schema-valid response rate decreases after $\varepsilon{\approx}0.6$,
falling from $100\%$ to $54.4\%$ and then to $3.0\%$ by the end of the
epoch. Full~FT shows a smaller PA decline and keeps schema validity at
or above $96\%$ throughout the same interval. Thus, the late-stage LoRA
drop is better interpreted as response-schema drift rather than lost
localization, and the paired benchmark makes this distinction measurable.

The trajectory also reveals a useful separation between paired-decision
learning and response-schema stability. The lightweight format term acts
as a schema regularizer, while the larger decision rewards drive the
detect-or-abstain behavior. After the paired decision has largely
converged, continued LoRA updates can therefore reveal schema drift even
when the underlying localization signal remains recoverable. Manual
inspection supports this interpretation: many late LoRA responses still
contain plausible boxes, but omit the opening \texttt{<think>} tag, for
example producing a closing \texttt{</think>} tag before an otherwise
recoverable bounding-box decision. Others deviate from the required tag
structure in similar ways. This observation motivates using schema
validity as a trajectory diagnostic when selecting the half-epoch
checkpoint, alongside PA and standard mask quality.

\subsection{General Multimodal Benchmarks}
\label{sec:analysis:capability}

\begin{figure}[!t]
  \centering
  \includegraphics[width=\columnwidth]{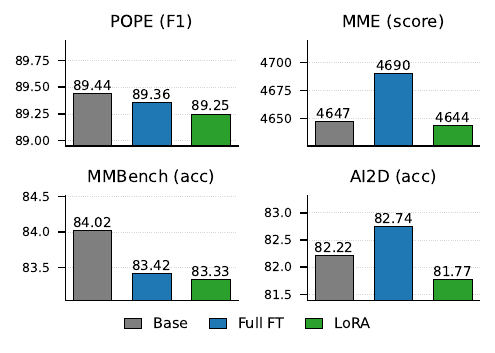}
  \caption{\normalfont Absolute scores on four multimodal
  benchmarks for the Qwen3-VL-4B-Instruct base and both \method{}
  variants. Note the zoomed vertical axes: across all four benchmarks
  every variant stays within $1\%$ (relative) of the base model, so
  the differences are marginal.}
  \label{fig:capability}
\end{figure}

Figure~\ref{fig:capability} asks whether the COD-specific RL stage
introduces a measurable loss on general multimodal tasks. We evaluate
\method{} on four multimodal benchmarks:
POPE~\cite{li2023EvaluatingObjectHallucination},
MME~\cite{fu2025MMEComprehensiveEvaluation},
MMBench~\cite{liu2024MMBenchYourMultimodal}, and
AI2D~\cite{kembhavi2016DiagramWorthDozen}, all under
VLMEvalKit~\cite{duan2024VLMEvalKitOpenSourceToolKit}. The comparison
uses the same vLLM stack and the same benchmark-specific prompt for all
models, without exposing the \method{} detect-or-abstain schema. These
benchmarks therefore serve as a regression check for general perception,
reasoning, multiple-choice question answering, and object-hallucination
behavior.

Both \method{} variants remain close to the Qwen3-VL-4B-Instruct base.
On the percentage-based metrics, the largest change is the LoRA model's
$0.69$~pp drop on MMBench; POPE and AI2D vary by at most $0.19$ and
$0.52$~pp, respectively. On MME, Full~FT improves by $42.6$ points
($0.9\%$ relative), while LoRA changes by less than $0.1\%$ relative to
the base. Thus, the paired COD improvements in Sections~\ref{sec:results}
and~\ref{sec:analysis:reward} do not come with an observable degradation
on these multimodal checks.

\section{Conclusion}
\label{sec:conclusion}

This study introduces \method{} and revisits camouflaged object detection
as a paired counterfactual problem. We argue that positive-only COD
training leaves target-absent behavior untested, even though ordinary
target-absent scenes are common deployment inputs. To make this behavior measurable, we
construct \benchmark{}, a paired detect-or-abstain benchmark where each
held-out COD image is paired with a target-absent counterpart synthesized by
an off-the-shelf inpainter. To reduce over-detection, CSPO uses paired counterfactual
rollouts, a paired sequence ratio extending GSPO, and CPR, which couples
original-image detection with counterfactual abstention through detection
and abstention indicators plus a pair-level coupling bonus.

Empirically, \method{} improves standard COD performance, strengthens paired
decisions on \benchmark{}, and shows no observable degradation on the
evaluated multimodal benchmarks.

Future work can further study robustness to the quality of synthesized
target-absent images, extend the same paired formulation to video COD
with temporal counterfactuals, and explore broader grounding tasks where
target existence is uncertain.

\appendices

\section{Prompt Templates}
\label{app:prompts}

We use two prompt templates. The detect-or-abstain prompt $P_{\mathrm{da}}$
is used for all RL training and for the \benchmark{} paired evaluation.
The standard prompt $P_{\mathrm{std}}$ is used only for the prior-art
comparison in Table~\ref{tab:std_cod}, where existing methods cannot
abstain. Both templates use Qwen3-VL's native $[0,1000]$ coordinate
normalization~\cite{bai2025Qwen3VLTechnicalReport}. The two templates
differ only in the \texttt{<no\_camouflage/>} option and the assertion
of target presence in $P_{\mathrm{std}}$.

\noindent\textbf{$P_{\mathrm{da}}$ — system message.}
\begin{lstlisting}[style=promptstyle]
You are a camouflaged object detector. Output in this exact format:
<think>your reasoning here</think>
followed by ONE of:
  - <bbox>[x1,y1,x2,y2]</bbox>           for a single object
  - <bbox>[[x1,y1,x2,y2],[x3,y3,x4,y4]]</bbox>  for multiple
  - <no_camouflage/>                      if none is present
Coordinates are normalized to [0, 1000] where 1000 = full image dimension.
\end{lstlisting}

\noindent\textbf{$P_{\mathrm{da}}$ — user message.}
\begin{lstlisting}[style=promptstyle]
Identify and locate any camouflaged object in the image.
In <think></think>, briefly consider scene textures, visual anomalies, and if any object blends in. Then output ONE of:
- <bbox>[x1,y1,x2,y2]</bbox> for one object, or [[x1,y1,x2,y2],...] for multiple
- <no_camouflage/> if no camouflaged object
\end{lstlisting}

\noindent\textbf{$P_{\mathrm{std}}$ — system message.}
\begin{lstlisting}[style=promptstyle]
You are a camouflaged object detector. There IS a camouflaged object in this image. Locate it precisely.
Output in this exact format:
<think>your reasoning here</think>
followed by ONE of:
  - <bbox>[x1,y1,x2,y2]</bbox>           for a single object
  - <bbox>[[x1,y1,x2,y2],[x3,y3,x4,y4]]</bbox>  for multiple
Coordinates are normalized to [0, 1000] where 1000 = full image dimension.
\end{lstlisting}

\noindent\textbf{$P_{\mathrm{std}}$ — user message.}
\begin{lstlisting}[style=promptstyle]
Identify and locate the camouflaged object in the image.
In <think></think>, briefly consider scene textures and visual anomalies, then output:
- <bbox>[x1,y1,x2,y2]</bbox> for one object, or [[x1,y1,x2,y2],...] for multiple
\end{lstlisting}

\section{SFT Ratio Sweep Data}
\label{app:sft_ratio_data}

\begin{table}[!h]
\caption{\normalfont Data underlying Fig.~\ref{fig:sft_ratio}: SFT ratio
sweep on CAMO-test (250 paired images, $P_{\mathrm{da}}$, $T{=}0$ greedy).
$D_o$, $A_c$, and PA are paired-decision rates (\%); $\mathrm{Fmt}$ is
the format-validity rate (\%), per the indicator defined in
Section~\ref{sec:method:reward}.
The 1:1 row is the balanced SFT used as cold-start for \method{} in the
main paper.}
\label{tab:sft_ratio_data}
\centering\footnotesize
\setlength{\tabcolsep}{6pt}
\begin{tabular}{lcccc}
\toprule
Ratio (det\,:\,abs) & $D_o\!\uparrow$ & $A_c\!\uparrow$ & $\mathrm{PA}\!\uparrow$ & $\mathrm{Fmt}\!\uparrow$ \\
\midrule
3\,:\,1 (detect-heavy)  & 90.4 & 49.2 & 42.8 & 98.4 \\
2\,:\,1                 & 87.6 & 57.6 & 46.4 & 99.4 \\
\rowcolor{gray!15}
\textbf{1\,:\,1 (balanced)} & 70.0 & 87.2 & \textbf{59.2} & 100.0 \\
1\,:\,2                 & 18.4 & 99.6 & 18.4 & 100.0 \\
1\,:\,3 (abstain-heavy) & 16.4 & 100.0 & 16.4 & 99.8 \\
\bottomrule
\end{tabular}
\end{table}

\end{document}